\documentclass[runningheads]{llncs}

\usepackage{eccv}

\usepackage{eccvabbrv}

\usepackage{graphicx}
\usepackage{booktabs}
\usepackage{xcolor}
\usepackage{multirow}
\usepackage[accsupp]{axessibility}  

\usepackage{hyperref}

\usepackage{orcidlink}

\begin{document}

\title{Mind2Cloud: EEG-to-Point Cloud Generation with Two-Granularity Diffusion Decoding} 

\titlerunning{Mind2Cloud}

\author{Yongyi Lu\inst{1}\orcidlink{0000-0003-1398-9965} \and
Xiongfeng Huang\inst{1}\orcidlink{0009-0008-7393-7482} \and
Zhijing Yang\inst{1}$^*$\orcidlink{0000-0001-8336-5109}}

\authorrunning{Y. Lu et al.}

\institute{Guangdong University of Technology, Guangzhou, China \\
\email{\{yylu, yzhj\}@gdut.edu.cn, huangxiongfeng@mails.gdut.edu.cn}}

\maketitle

\let\thefootnote\relax\footnotetext{$^*$ Corresponding author.}

\begin{abstract}
  Reconstructing 3D objects from brain signals offers a promising avenue for understanding human visual cognition. While prior work has shown initial success using EEG signals for 3D reconstruction, existing methods typically employ a uniform diffusion decoder, overlooking the evolving semantic granularity of both EEG representations and the diffusion denoising process. In this paper, we propose Mind2Cloud, a novel EEG-to-point-cloud generation framework based on two-granularity diffusion decoding. The core of Mind2Cloud is a time-aware decoder that integrates a global Transformer branch and a local Point-Voxel CNN (PVCNN) branch across diffusion timesteps through a learnable fusion mask. Specifically, Transformer layers are incorporated into the early upsampling stages to capture global object structure under high uncertainty, while PVCNN modules are used in later stages to refine local geometric details. Inspired by the hierarchical nature of EEG-based visual representations, this design dynamically adapts its spatial granularity in accordance with the coarse-to-fine trajectory of diffusion denoising. We further introduce an adversarial refinement module to enhance geometric realism and semantic consistency. Extensive experiments on the EEG-3D dataset across all 12 subjects demonstrate that Mind2Cloud outperforms prior work in both geometric accuracy and semantic alignment, setting a new benchmark for EEG-to-point-cloud generation. Our source code is available at \url{https://github.com/duasoi/Mind2Cloud}.

\end{abstract}

\section{Introduction}
\label{sec:intro}

Reconstructing 3D shapes from non-invasive brain signals is a fundamental challenge in neural decoding, as it requires mapping noisy, low-dimensional neural activity to structured, high-fidelity geometry representations. Prior studies have made notable progress in reconstructing 2D images from fMRI or EEG signals~\cite{lin2022mind, scotti2023reconstructing, chen2023cinematic}. However, extending visual decoding from 2D images to 3D is substantially more challenging, due to the larger domain gap between neural signals and geometric structures. Recent work such as Neuro-3D~\cite{guo2025neuro} has demonstrated the feasibility of EEG-to-3D shape reconstruction by combining an EEG encoder for static and dynamic signal learning with a diffusion-based point cloud generator. Nevertheless, existing methods typically employ a fixed decoding architecture throughout the entire generation process, overlooking the fact that the semantic granularity required for reconstruction evolves across both EEG representations and the diffusion denoising process.

\begin{figure}[t]
\centering
\includegraphics[width=0.7\columnwidth]{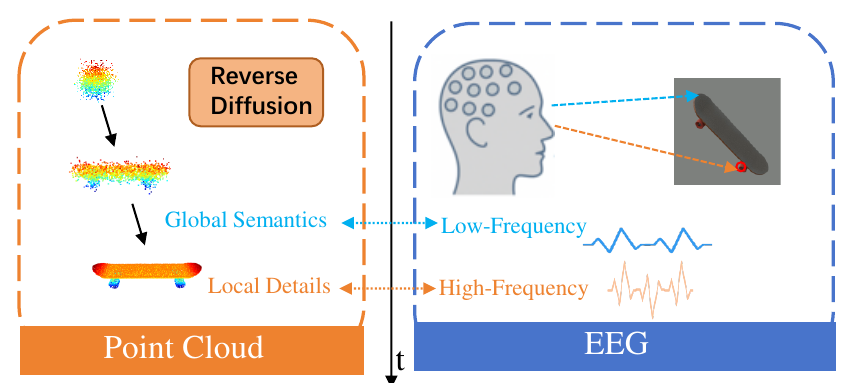} 
\caption{Conceptual illustration of two-granularity diffusion decoding for EEG-to-point-cloud generation. Low-frequency EEG cues guide global semantic structure during early denoising, while high-frequency cues support local detail refinement in later steps.}
\label{fig:teaser}
\end{figure}

Neuroscientific evidence suggests that EEG carries visual information at multiple levels of abstraction: low-frequency components are often associated with global object-level perception, while high-frequency components are more closely related to local visual details such as edges, contours and fine structures~\cite{cichy2016deep, zhang2025cognitioncapturer}. A similar hierarchy naturally appears in diffusion-based point cloud generation. During early denoising steps, the model operates on highly noisy latent shapes and must recover coarse global structure. In later steps, it progressively refines local geometry and details, as illustrated in~\ref{fig:teaser}. This conceptual hierarchy serves as a key inspiration for designing a 3D visual decoding architecture from EEG that can adapt its spatial granularity over time, in alignment with the coarse-to-fine trajectory of diffusion generation.

To this end, we propose \textit{\textbf{Mind2Cloud}}, a two-granularity diffusion decoder for EEG-guided point cloud reconstruction. Our decoder integrates a global Transformer branch and a local Point-Voxel CNN (PVCNN) branch, fusing their outputs in a timestep-aware manner. Specifically, during the diffusion process, we embed Transformer blocks into the early upsampling stages of the decoder, where the model must synthesize coarse structure under high uncertainty. As denoising progresses and the generated point cloud becomes more structured, we gradually reduce the influence of the Transformer and rely on the PVCNN branch to refine spatial detail. A learnable fusion mask modulates the relative contribution of each branch at different timesteps, allowing the network to smoothly transition from global semantic reconstruction to local geometric refinement. Rather than a mere architectural combination, the key novelty of Mind2Cloud lies in its EEG-conditioned, timestep-adaptive global/local fusion mechanism. This design dynamically balances complementary decoding granularity across denoising stages and implicitly aligns the hierarchical information encoded in EEG signals with the generative process.

To further enhance geometric realism, we introduce a lightweight adversarial refinement module that discriminates between reconstructed and real point clouds. The entire model is trained with a combination of contrastive alignment loss, geometric reconstruction loss and adversarial supervision. We validate Mind2Cloud on the EEG-3D dataset across all 12 subjects, demonstrating substantial improvements over prior work in both perceptual quality and reconstruction accuracy. The proposed decoder produces more structurally faithful 3D shapes, better preserves semantic correspondence with the neural signals without increasing point cloud resolution or inference cost.

In summary, our main contributions are as follows:
\begin{itemize}
    \item We show that selectively incorporating Transformer modules into early upsampling stages improves EEG-to-point-cloud generation by better matching the coarse-to-fine nature of both neural visual perception and diffusion decoding.
    \item We introduce a two-granularity point cloud decoder featuring an EEG-conditioned, timestep-adaptive fusion mechanism that adaptively fuses global Transformer and local PVCNN features across diffusion timesteps.
    \item Mind2Cloud sets a new benchmark for EEG-to-point-cloud generation on the EEG-3D dataset under full-subject evaluation across all 12 subjects, achieving improved geometric accuracy and semantic fidelity.
\end{itemize}

\section{Related Work}
\subsection{Neural Decoding of 2D Visual Stimuli}
Decoding visual stimuli from neural signals ~\cite{chen2024eegformer,chen2023cinematic,chen2023seeing,lahner2024modeling,scotti2023reconstructing,zhang2025cognitioncapturer} has emerged as a key research direction bridging computer vision and neuroscience, aiming to reveal how the brain encodes visual information. Early efforts leveraged CNNs and GANs to map fMRI or EEG signals to 2D images~\cite{beliy2019voxels,horikawa2017generic,lin2022mind,ozcelik2022reconstruction,shen2019deep,goodfellow2020generative}, demonstrating the viability of reconstructing perceptually coherent visuals from brain activity. Recent advances in diffusion models~\cite{ho2020denoising,nichol2021improved} and vision-language frameworks such as CLIP~\cite{li2023blip} have further enhanced decoding fidelity. Many approaches align neural representations with pre-trained visual or textual embeddings via contrastive learning, enabling conditional generation from brain signals~\cite{wang2020understanding,radford2021learning,bao2025wills}. Despite success in static 2D and short video reconstruction~\cite{chen2023cinematic,kupershmidt2022penny,sun2024neurocine}, these methods remain constrained to 2D, limiting their ability to capture the spatial depth of 3D perception.

\subsection{3D Reconstruction from Brain Signals}

Reconstructing 3D objects from neural activity is an emerging direction in neural decoding, with promising applications in brain–computer interfaces (BCIs) and understanding spatial perception. Early work mainly used fMRI due to its high spatial resolution to recover coarse 3D structures~\cite{gao2024fmri,gao2024mind}. Projects like Mind-3D~\cite{gao2024mind} and Mind-3D++~\cite{gao2024mind++} introduced paired datasets and diffusion-based generative models for geometry inference. However, fMRI methods are costly, immobile, and have poor temporal resolution~\cite{Polimeni2010}, limiting real-time or large-scale BCI use. They also often miss perceptual details like color and texture, important for semantic fidelity. To address this, Neuro-3D~\cite{guo2025neuro} proposed the first EEG-based 3D reconstruction dataset and framework, allowing geometry and appearance estimation from non-invasive EEG signals. EEG is affordable, portable, and temporally responsive, better suited for interactive real-time BCIs~\cite{willett2021high,jin2024pgcn}. Despite this, EEG-to-3D reconstruction accuracy and completeness remain limited, especially under noisy, low-SNR conditions. This motivates further research on improving neural representations, model robustness, and cross-modal alignment to advance reliable, semantically rich 3D brain decoding systems.

\subsection{Diffusion Models}
Diffusion models have become a powerful generative paradigm capable of synthesizing high-quality visual content by reversing a noise corruption process~\cite{ho2020denoising}. Their applications span 2D image generation \cite{kim2022diffusionclip,ramesh2021zero,rombach2022high,saharia2022photorealistic}, editing \cite{yu2024accelerating,zhang2023adding}, and 3D tasks including point cloud modeling \cite{vahdat2022lion,zhou20213d}, multi-view synthesis \cite{long2024wonder3d,melas2023pc2,hu20242,zou2024sparse3d}, and text-to-3D reconstruction \cite{nichol2022point,poole2022dreamfusion,chen2024it3d,li2024focaldreamer}. To enhance efficiency and detail fidelity, latent diffusion models compress inputs via autoencoders \cite{rombach2022high}, while Transformer-based architectures like DiT improve scalability and generation quality \cite{peebles2023scalable}. In neuroscience, diffusion models have been leveraged for fMRI-based visual decoding, with frameworks like TIGER \cite{ren2024tiger} combining convolutional and Transformer modules to reconstruct dynamic 3D shapes. Building on these insights, we extend diffusion-based generation to EEG-driven 3D reconstruction, establishing a new direction for neural-guided generative modeling.

\section{Methodology}
\subsection{Overview}

\begin{figure*}[!t]
  \centering
  \includegraphics[width=\textwidth]{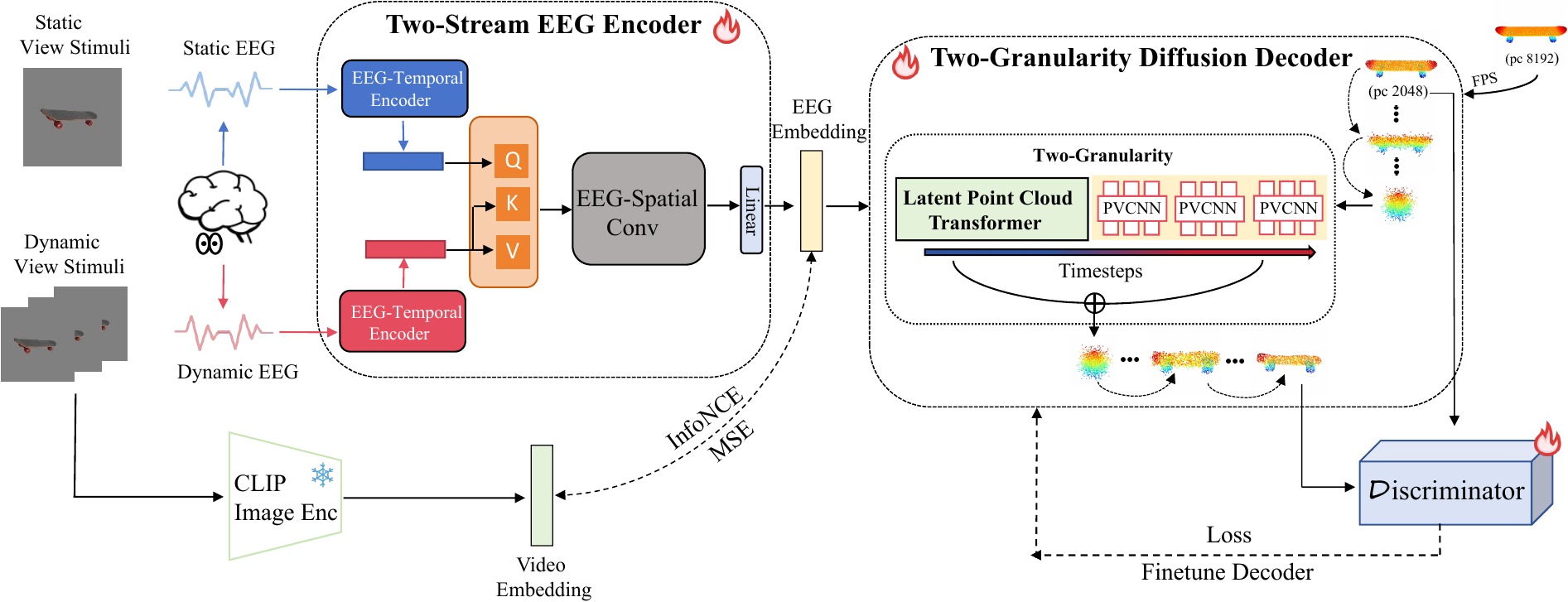}
  \caption{Overview of Mind2Cloud. A two-stream EEG encoder extracts complementary features from static and dynamic EEG responses and aligns them with CLIP image embeddings. The resulting EEG embedding guides a two-granularity diffusion decoder to reconstruct 3D point clouds, while adversarial refinement improves geometric fidelity.}
  \label{method}
\end{figure*}

We propose an end-to-end framework for reconstructing 3D shapes from EEG signals, as illustrated in Figure \ref{method}. The framework consists of three main components: 1)\textbf{ Two-stream EEG Encoder}, which adaptively integrates static and dynamic EEG signals to extract complementary neural representations. Motivated by the Neuro-3D~\cite{guo2025neuro} paradigm, this design enhances the expressiveness of EEG features for 3D object understanding. To ensure semantic alignment across modalities, we introduce InfoNCE and MSE losses between the EEG embeddings and CLIP-derived visual embeddings; 2)\textbf{ Point Cloud Decoder}, which reconstructs 3D shapes conditioned on the EEG embedding by leveraging a latent point cloud Transformer. This decoder fuses global representations from the Transformer and local geometric features extracted via shallow PVCNN modules~\cite{liu2019point} using a timestep-aware weighting strategy; and 3)\textbf{ Adversarial Refinement}, where a discriminator is employed to distinguish generated point clouds from ground-truth samples. This adversarial objective fine-tunes the decoder, guiding it to produce more realistic and structurally coherent 3D outputs.

\subsection{Two-stream EEG Encoder}

To extract robust neural representations from temporally and spatially distributed EEG signals, we propose a dual-branch EEG encoder that separately processes static and dynamic modalities and integrates them via temporal attention and spatial convolution.

\subsubsection{EEG-Temporal Encoder.}
Given preprocessed EEG signals $x_s \in \mathbb{R}^{C \times T_s}$ and $x_d \in \mathbb{R}^{C \times T_d}$ corresponding to static and dynamic visual stimuli, we employ two Transformer encoders to model their temporal structures. Each input sequence is enriched with learnable positional encodings and encoded as:
\begin{equation}
\begin{split}
\mathbf{z}_{\text{dyn}} &= \text{Transformer}(\text{PosEnc}(x_d)) \\
\mathbf{z}_{\text{stc}} &= \text{Transformer}(\text{PosEnc}(x_s))
\end{split}
\end{equation}

To fuse temporal cues from both branches, we apply a cross-attention mechanism where static features serve as the query and dynamic features as key–value pairs:
\begin{equation}
\mathbf{z}_{\text{fuse}} = \text{CrossAttn}(Q = \mathbf{z}_{\text{stc}},\ K = \mathbf{z}_{\text{dyn}},\ V = \mathbf{z}_{\text{dyn}}),
\end{equation}
yielding a unified temporal representation that integrates static and motion-related information.

\subsubsection{EEG-Spatial Conv.}
\label{sec:eeg_spatial}
To model inter-channel dependencies, we apply a series of 1D convolutional layers with residual connections to $\mathbf{z}_{\text{fuse}}$, followed by a linear projection to generate the final EEG embedding. This spatial encoding enhances the topographical structure of EEG signals and stabilizes training.

To align EEG and visual representations, we adopt a symmetric InfoNCE loss:
\begin{equation}
\mathcal{L}_{\text{clip}} =
\frac{1}{2}
\left[
\mathrm{CE}\left(\frac{\mathbf{z}_{\text{fuse}} \mathbf{f}_v^\top}{\tau}, y\right)
+
\mathrm{CE}\left(\frac{\mathbf{f}_v \mathbf{z}_{\text{fuse}}^\top}{\tau}, y\right)
\right],
\end{equation}
where $\tau$ is a temperature parameter, and an additional regression term for direct embedding alignment:
\begin{equation}
\mathcal{L}_{\text{img}} = \left\| \mathbf{z}_{\text{fuse}} - \mathbf{f}_v \right\|_2^2.
\end{equation}

The combined loss is defined as:
\begin{equation}
\mathcal{L}_{\text{EEG}} = \mathcal{L}_{\text{img}} + \mathcal{L}_{\text{clip}},
\end{equation}
encouraging EEG embeddings that are both semantically aligned and discriminative for downstream 3D generation.

\subsection{Two-Granularity Diffusion Decoder}
To reconstruct high-fidelity 3D shapes from noisy EEG-derived embeddings, we design a \textbf{time-aware Two-Granularity point cloud decoder} that adaptively adjusts its spatial granularity across diffusion timesteps. While Transformer and CNN architectures are standard modules, their complementary inductive biases remain highly effective for structured 3D generation tasks. In our design, the Transformer branch captures long-range dependencies and global semantic structure, whereas the CNN-based PVCNN branch focuses on localized geometric refinement and high-frequency spatial details. Importantly, the innovation of our framework lies not in architectural reinvention, but in the EEG-conditioned diffusion step–adaptive fusion mechanism. This mechanism dynamically balances global and local representations across denoising stages while being guided by neural embeddings extracted from brain activity. By coupling EEG semantics with timestep-aware feature modulation, the decoder progressively translates abstract neural representations into structured geometric details, achieving semantically grounded and structurally coherent 3D generation. This design is motivated by the coarse-to-fine trajectory of diffusion-based generation and the hierarchical abstraction levels present in EEG signals. Drawing inspiration from the TIGER architecture~\cite{ren2024tiger}, our decoder integrates a local CNN-based branch for geometric refinement, a Transformer-based branch for modeling global structure, and a learnable fusion mechanism that dynamically balances the two throughout the denoising process. This enables the decoder to progressively transition from global shape synthesis to fine-grained spatial detail, aligning feature representation with the generative stage.

\subsubsection{Local Geometric Refinement.}
Given a noisy point cloud $\mathbf{X}_t \in \mathbb{R}^{N \times 3}$ at diffusion timestep $t$, we first concatenate it with EEG-derived latent embeddings to form a joint representation. To extract high-frequency spatial details and local semantic cues, we adopt a hierarchical feature extractor built upon the Point-Voxel Convolutional Neural Network (PVCNN) framework. 

Specifically, we employ a multi-stage Set Abstraction (SA) architecture that combines point-based and voxel-based processing. In each stage, point-wise MLPs capture fine-grained variations by aggregating features from local $k$-NN neighborhoods, while voxel-based convolutions introduce geometric regularity and reduce sensitivity to spatial noise. This hybrid design allows the model to simultaneously leverage unstructured geometric richness and structured volumetric priors, making it well-suited for EEG-guided generation where signal uncertainty is high.

The extracted local features are progressively downsampled and aggregated, preserving intermediate coordinates and activations for later upsampling via Feature Propagation (FP) layers. This enables effective gradient flow and spatial correspondence between encoder and decoder stages. Overall, the PVCNN branch focuses on capturing localized structures—such as edges, surface curvature, and fine object parts—that are critical for reconstructing semantically meaningful and geometrically detailed 3D shapes. When integrated with the Transformer-based global branch, it provides complementary inductive bias that enhances both structural fidelity and robustness during generation.

\subsubsection{Global Semantic Reconstruction.}
To effectively model global spatial relationships within latent point clouds, we introduce a Transformer-based module operating at an intermediate resolution. Specifically, given latent point cloud features $\hat{\mathbf{X}}_t \in \mathbb{R}^{M \times d}$, we first encode them into initial token embeddings using a dual normalization~\cite{kumar2023dual} and intermediate MLP scheme:
\begin{equation}
\mathbf{T}_0 = \mathrm{LN}\bigl(\mathrm{MLP}(\mathrm{LN}(\hat{\mathbf{X}}_t))\bigr) \in \mathbb{R}^{M \times D}.
\end{equation}

\begin{figure}[t]
\centering
\includegraphics[width=0.7\columnwidth]{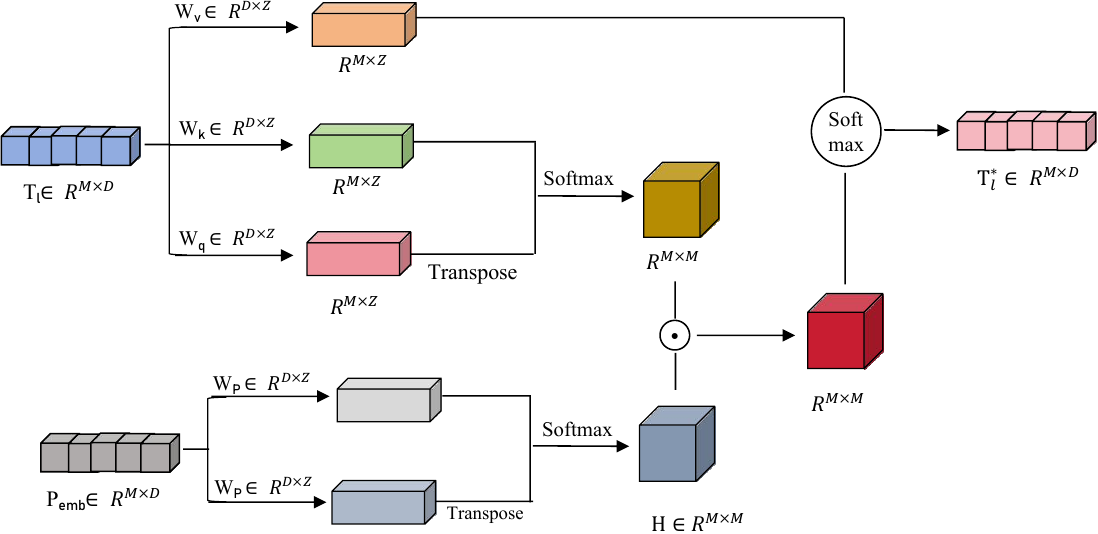} 
\caption{Position-aware attention mechanism with a learned spatial interaction matrix $H$, integrating positional embeddings into Transformer attention for enhanced 3D structural modeling.}
\label{fig:pos}
\end{figure}

We further enrich these embeddings with explicit positional information by applying a continuous, multi-granularity 3D positional embedding~\cite{ren2024tiger} $\text{Pemb} \in \mathbb{R}^{M \times D}$. This encoding scheme effectively captures fine-grained spatial differences and maintains continuity across spatial dimensions, essential for modeling irregular and continuous 3D data. Inspired by~\cite{zhao2021point}, rather than directly adding positional embeddings to token representations, which could diminish positional information across deeper layers, we integrate spatial context through a learned spatial interaction matrix (Fig.~\ref{fig:pos}):
\begin{equation}
H = \mathrm{Softmax}\left((\text{Pemb}W_p)(\text{Pemb}W_p)^\top\right),
\end{equation}
where $W_p \in \mathbb{R}^{D \times Z}$ is a projection matrix mapping the positional embeddings into an attention-specific space. This learned interaction explicitly preserves positional relationships, enabling consistent incorporation of relative spatial structure throughout successive Transformer layers.

Formally, we stack $L$ Transformer layers, where at each layer $l \in \{0,1,...,L-1\}$, the token embedding $\mathbf{T}_l \in \mathbb{R}^{M \times D}$ is iteratively refined. The position-aware attention computation within each Transformer layer is defined as:
\begin{equation}
\mathbf{T}_l^* = \mathrm{Softmax}\left(\frac{(\mathbf{T}_l W_q)(\mathbf{T}_l W_k)^\top \odot H}{\sqrt{Z}}\right)(\mathbf{T}_l W_v),
\end{equation}

where $W_q, W_k, W_v \in \mathbb{R}^{D \times Z}$ are the query, key, and value projection matrices, and $Z$ denotes the dimensionality of the projected query and key vectors. A feed-forward network follows each attention layer to further refine the token embeddings. After $L$ iterations, we obtain the final representation $\mathbf{F}_{\text{tr}}$, which encodes high-level semantic structure and serves as a global prior for downstream reconstruction.

This Transformer-based branch is particularly effective during early diffusion steps, where global semantics dominate and spatial uncertainty remains high. By capturing object-level structure and integrating geometric context over long distances, it complements the local refinement pathway and enhances the fidelity of 3D shape reconstruction.

\subsubsection{Time-aware Feature Fusion.}
To dynamically integrate coarse global structure with fine local details across the generative trajectory, we design a \textbf{time-aware feature fusion module} that adaptively modulates the contributions of the Transformer and CNN branches throughout the diffusion process. This design reflects the observation that early timesteps favor semantic abstraction, while later stages benefit from localized refinement.

Specifically, the diffusion timestep $t$ is encoded using sinusoidal positional embeddings and fed into an MLP to generate a learnable fusion mask $M_t \in \mathbb{R}^{D}$. The mask is constrained to $(0,1)$ via a sigmoid activation and applied channel-wise, with broadcasting along spatial dimensions.

The fused feature is computed as:
\begin{equation}
F_{\text{out}} =
M_t \cdot \text{Conv}(F_{\text{conv}})
+
(1 - M_t) \cdot \text{TF}(F_{\text{tr}}),
\end{equation}
where $\text{Conv}(\cdot)$ and $\text{TF}(\cdot)$ denote $1 \times 1$ learnable convolutions that project both branches into a shared latent space. The mask $M_t$ allows the network to emphasize global structural information during early denoising stages (high noise), and progressively shift toward local detail enhancement as uncertainty decreases over time.

The fused representation $F_{\text{out}}$ is further processed by normalization and residual blocks, followed by a point-wise MLP that predicts the noise residual $\hat{\epsilon}$ for each 3D point. This output is used to guide the denoising process, refining the noisy point cloud $\mathbf{X}_t$ toward the target shape $\mathbf{X}_0$. The diffusion reconstruction is supervised via a standard noise prediction loss:
\begin{equation}
\mathcal{L}_{\text{diffuse}} = \|\hat{\epsilon} - \epsilon\|_2^2,
\end{equation}
where $\epsilon$ is the Gaussian noise added during the forward diffusion step.

To jointly optimize geometric accuracy and EEG-image alignment, we combine this diffusion loss with the EEG supervision loss $\mathcal{L}_{\text{EEG}}$ (as introduced in Section~\ref{sec:eeg_spatial}), which includes an InfoNCE contrastive term $\mathcal{L}_{\text{clip}}$ and a regression term $\mathcal{L}_{\text{img}}$. The overall training objective is defined as:
\begin{equation}
\mathcal{L}_{\text{joint}} = \alpha \cdot \mathcal{L}_{\text{diffuse}} + (1 - \alpha) \cdot \mathcal{L}_{\text{EEG}},
\label{L}
\end{equation}
where $\alpha \in [0,1]$ trades off between accurate shape reconstruction and cross-modal representation learning.

\begin{figure}[t]
\centering
\includegraphics[width=0.7\textwidth]
{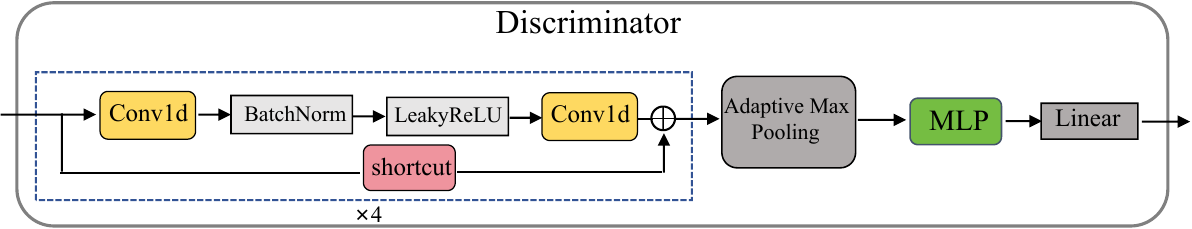} 
\caption{Discriminator architecture. The input point cloud is processed by four stacked residual Conv1D blocks, followed by global pooling and an MLP classifier.}
\label{D}
\end{figure}

\subsection{Adversarial Refinement}
To enhance the geometric realism of reconstructed shapes, we introduce a lightweight point cloud discriminator that distinguishes real samples from generated ones. Given an input point cloud $\mathbf{X} \in \mathbb{R}^{N \times 3}$, the discriminator outputs a scalar authenticity score.

As shown in Figure~\ref{D}, the discriminator leverages stacked residual 1D convolutional blocks to model local geometry and channel dependencies. A global max pooling followed by normalization produces a compact representation, which is mapped to a binary prediction.

During training, adversarial learning is applied using binary cross-entropy loss on logits (BCEWithLogitsLoss). Given real samples $\mathbf{X}_0$ and generated ones $\tilde{\mathbf{X}}$, the discriminator loss is:
\begin{equation}
\mathcal{L}_{\text{D}} = \text{BCE}(D(\mathbf{X}_0), 1) + \text{BCE}(D(\tilde{\mathbf{X}}), 0),
\end{equation}
where $D(\cdot)$ denotes the discriminator output before activation.


Meanwhile, the generator is trained with an adversarial objective that encourages indistinguishability from real shapes:
\begin{equation}
\mathcal{L}_{\text{G}} = \text{BCE}(D(\tilde{\mathbf{X}}), 1).
\end{equation}

This adversarial term is combined with the primary loss $\mathcal{L}$—which includes diffusion-based reconstruction and EEG alignment—yielding the final training objective:
\begin{equation}
\mathcal{L}_{\text{total}} = \mathcal{L}_{\text{joint}} + \lambda_{\text{adv}} \cdot \mathcal{L}_{\text{G}},
\label{total}
\end{equation}
where $\lambda_{\text{adv}}$ balances the influence of adversarial feedback. This joint formulation promotes structurally consistent and perceptually realistic 3D outputs.

\section{Experiments}
\subsection{Datasets}
We adopt the same dataset as the Neuro-3D framework, a multimodal benchmark tailored for EEG-driven 3D reconstruction. The dataset comprises EEG recordings from 12 participants, each exposed to both static images and 360-degree rotating videos of 3D objects drawn from 72 categories in the Objaverse dataset~\cite{deitke2023objaverse,xu2024pointllm}. Each object is annotated with shape category, textual description, and color style labels. EEG signals were recorded using a 64-channel EASYCAP system at a sampling rate of 1000 Hz. A controlled stimulus paradigm was employed, presenting each object with a 0.5-second static image followed by a 6-second rotation video to elicit rich visual responses. Each participant completed approximately 5.5 hours of EEG acquisition, including resting-state, image-evoked, and video-evoked sessions. EEG preprocessing was conducted using the MNE toolbox. For more details, we refer readers to the original Neuro-3D study~\cite{guo2025neuro}.

\subsection{Implementation Details}
All experiments were performed on a single NVIDIA A6000 GPU, with total training time of approximately 2.5 days. We employed the AdamW optimizer with an initial learning rate of $1 \times 10^{-3}$. The visual encoder produced feature vectors of dimension 1024. In training, the loss weight $\alpha$ in Equation~\eqref{L} was set to 0.95 to emphasize precise geometric reconstruction, while the adversarial loss weight $\lambda_{\text{adv}}$  in Equation~\eqref{total} was set to 0.05 to incorporate moderate discriminator guidance for improving surface fidelity. The Point Cloud Decoder utilized an 8-layer Transformer architecture. To maintain manageable memory usage and computational requirements, we applied Farthest Point Sampling (FPS)~\cite{moenning2003fast} to downsample each raw point cloud from 8192 to 2048 points before training. Specifically, FPS is performed dynamically on the ground-truth meshes to ensure uniform spatial distribution. For semantic evaluation, since the original baseline classifier was not publicly released, we retrained our own PointNet++ classifier~\cite{qi2017pointnet++} following their evaluation protocol. To guarantee a fair evaluation, all test-set-related samples were strictly excluded during its training. This classifier was trained for 200 epochs using an Adam optimizer with a batch size of 32 and a learning rate of $1 \times 10^{-4}$ to ensure stable convergence.

\begin{figure*}[t]
\centering
\includegraphics[width=0.95\linewidth]{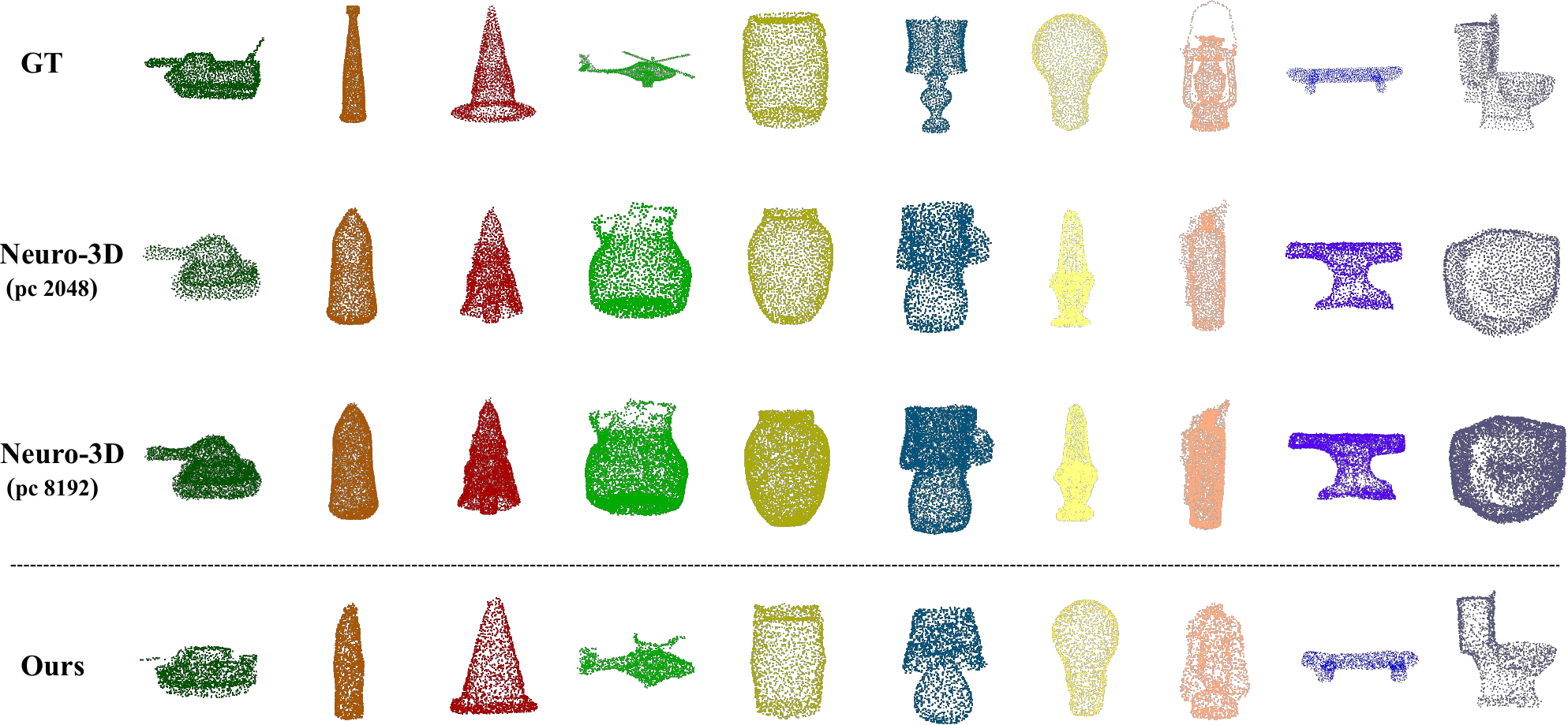} 
\caption{Qualitative comparison of reconstructed point clouds for Subject 1. Rows show ground-truth point clouds, FPS-downsampled Neuro-3D outputs with 2048 points, original Neuro-3D outputs with 8192 points, and our 2048-point reconstructions. Despite using fewer points, Mind2Cloud better preserves object structure and local geometry.}
\label{vis}
\end{figure*}

\begin{figure}[t]
    \centering
    \includegraphics[width=0.6\textwidth]{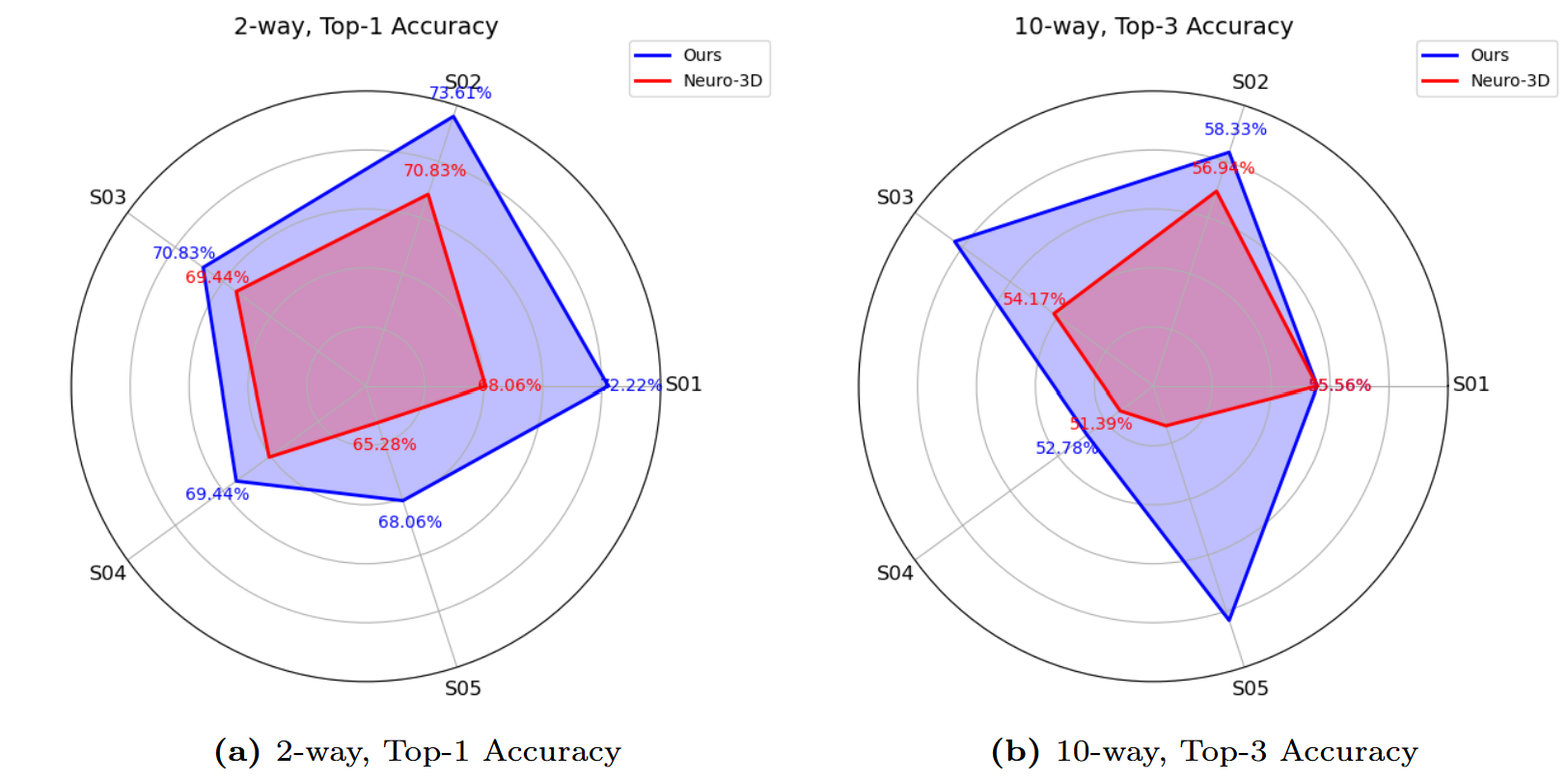}
    \caption{Subject-wise radar plots of the 2-way/top-1 accuracy and 10-way/top-3 settings for S01–S05. Mind2Cloud consistently outperforms Neuro-3D under both evaluations.}
    \label{fig:radar_subjectwise}
\end{figure}

\subsection{Results}
\subsubsection{Visualization Results.}
To qualitatively evaluate reconstruction performance, we visualize representative 3D point clouds from Subject~1 in Figure~\ref{vis}. The first row displays the ground-truth (GT) shapes, while the second and third rows show results from the Neuro-3D baseline, originally trained with 8192-point clouds. For fair comparison, we apply Farthest Point Sampling (FPS) to downsample these outputs to 2048 points, matching our input resolution. Both predicted shapes and corresponding ground-truth point clouds are uniformly resampled to 2048 points before metric computation to ensure fair comparison across methods. The last row presents results from our method trained and tested directly with 2048 points. All point clouds are rendered via Open3D with automatically assigned colors. Despite the reduced point density, our approach reconstructs finer geometric details with sharper contours and fewer deformations. In particular, in complex categories such as \textit{lightbulb}, \textit{helicopter}, and \textit{skateboard}, our model better preserves global structure and part connectivity. These results highlight the effectiveness of our framework in generating high-fidelity 3D shapes from EEG signals with improved efficiency.

\begin{table}[t]
\centering\tiny
\resizebox{0.7\textwidth}{!}{%
\begin{tabular}{c|l|c|c}
\toprule
\textbf{Datasets} & \textbf{Method} & \textbf{CD $\downarrow$} & \textbf{EMD $\downarrow$} \\
\midrule
\multirow{6}{*}{EEG}
& Neuro-3D~\cite{guo2025neuro} & 4.32 & 16.31 \\
& Ours (w/o Discriminator) & 3.26 & 15.93 \\
& Ours (w/o Two-Granularity) & 3.20 & 15.93 \\
& Ours (w/o PVCNN) & 3.71 & 17.10 \\
& Ours (Fixed Fusion) & 3.00 & 15.42 \\
& \textbf{Ours} & \textbf{2.53} & \textbf{14.02} \\
& Ours (S01-S12) & 2.65 & 14.43 \\
\bottomrule
\end{tabular}
}
\caption{Comparison of reconstruction fidelity using Chamfer Distance (CD) and Earth Mover’s Distance (EMD), both multiplied by $\times 10^2$. Lower values indicate better geometric accuracy.}
\label{tab:cd_emd}
\end{table}

\begin{table}[t]
\centering\tiny
\resizebox{.8\textwidth}{!}{
\begin{tabular}{l|cc|cc}
\toprule
\multirow{2}{*}{\textbf{Method}} & \multicolumn{2}{c|}{\textbf{Average}} & \multicolumn{2}{c}{\textbf{Top-1 of 5 samples}} \\
 & 2-way/top-1 & 10-way/top-3 & 2-way/top-1 & 10-way/top-3 \\
\midrule
Neuro-3D ~\cite{guo2025neuro} & 50.80\% & 31.72\% & 68.33\% & 53.89\% \\
\cmidrule(lr){1-5}
Ours (w/o Discriminator) & 50.65\% & 31.80\% & 66.94\% & 52.22\% \\
Ours (w/o Two-Granularity) & 50.12\% & 31.05\% & 67.22\% & 51.95\% \\
Ours (w/o PVCNN) & 49.58\% & 29.58\% & 50.56\% & 37.50\% \\
Ours (Fixed Fusion) & 50.69\% & 32.36\% & 66.39\% & 52.22\% \\
\textbf{Ours} & \textbf{52.92\%} & \textbf{32.47\%} & \textbf{70.83\%} & \textbf{56.67\%} \\
Ours (S01-S12) & 51.68\% & 31.30\% & 70.53\% & 52.84\% \\
\bottomrule
\end{tabular}
}
\caption{Semantic reconstruction accuracy under N-way Top-$k$ classification. Results are reported for the 2-way/top-1 and 10-way/top-3 settings, including average accuracy and Top-1 accuracy over five sampled candidates. Higher values indicate better semantic preservation.}
\label{tab:nway}
\end{table}

\subsubsection{Quantitative Evaluation.}
Since EEG-driven 3D point cloud reconstruction remains a relatively emerging research direction, publicly available and reproducible works are still limited. Among existing approaches, Neuro-3D is a representative and recently proposed method with publicly released models and code, making it suitable for direct and fair comparison. Therefore, we adopt Neuro-3D as our primary baseline and follow its evaluation protocol to ensure consistency and reproducibility. To quantitatively evaluate the geometric fidelity and semantic alignment of reconstructed 3D shapes, we report Chamfer Distance (CD) and Earth Mover’s Distance (EMD) in Table~\ref{tab:cd_emd}, and N-way Top-$k$ classification accuracy in Table~\ref{tab:nway}. We follow Neuro-3D and report results on the same five subjects to ensure direct comparability with the prior work. While the full dataset contains 12 participants, we adopt this evaluation protocol for fair benchmarking. Our method significantly outperforms the EEG-based baseline Neuro-3D on both CD and EMD, demonstrating superior reconstruction quality. 

Furthermore, to demonstrate cross-subject robustness, we extend our quantitative evaluation to the remaining subjects (S06–S12) and report the overall averages across all 12 subjects (S01–S12) at the bottom of Table~\ref{tab:cd_emd} and Table~\ref{tab:nway}. The stable and consistent performance across the entire participant pool demonstrates that our framework generalizes well to unseen subjects, successfully mitigating the inherent variability of neural signals.

For semantic evaluation, we retrain a PointNet++~\cite{qi2017pointnet++} classifier (84\% accuracy) since the original Neuro-3D classifier (90\%) is not publicly available. As shown in Table~\ref{tab:nway}, our method surpasses Neuro-3D in both 2-way and 10-way classification, indicating better semantic preservation. Ablation studies further verify the contribution of each component. Figure~\ref{fig:radar_subjectwise} visualizes per-subject accuracy of 2-way/top-1 and 10-way/top-3 settings, our method consistently outperforms Neuro-3D under both evaluations.

\begin{figure}[t]
    \centering
    \includegraphics[width=0.7\textwidth]{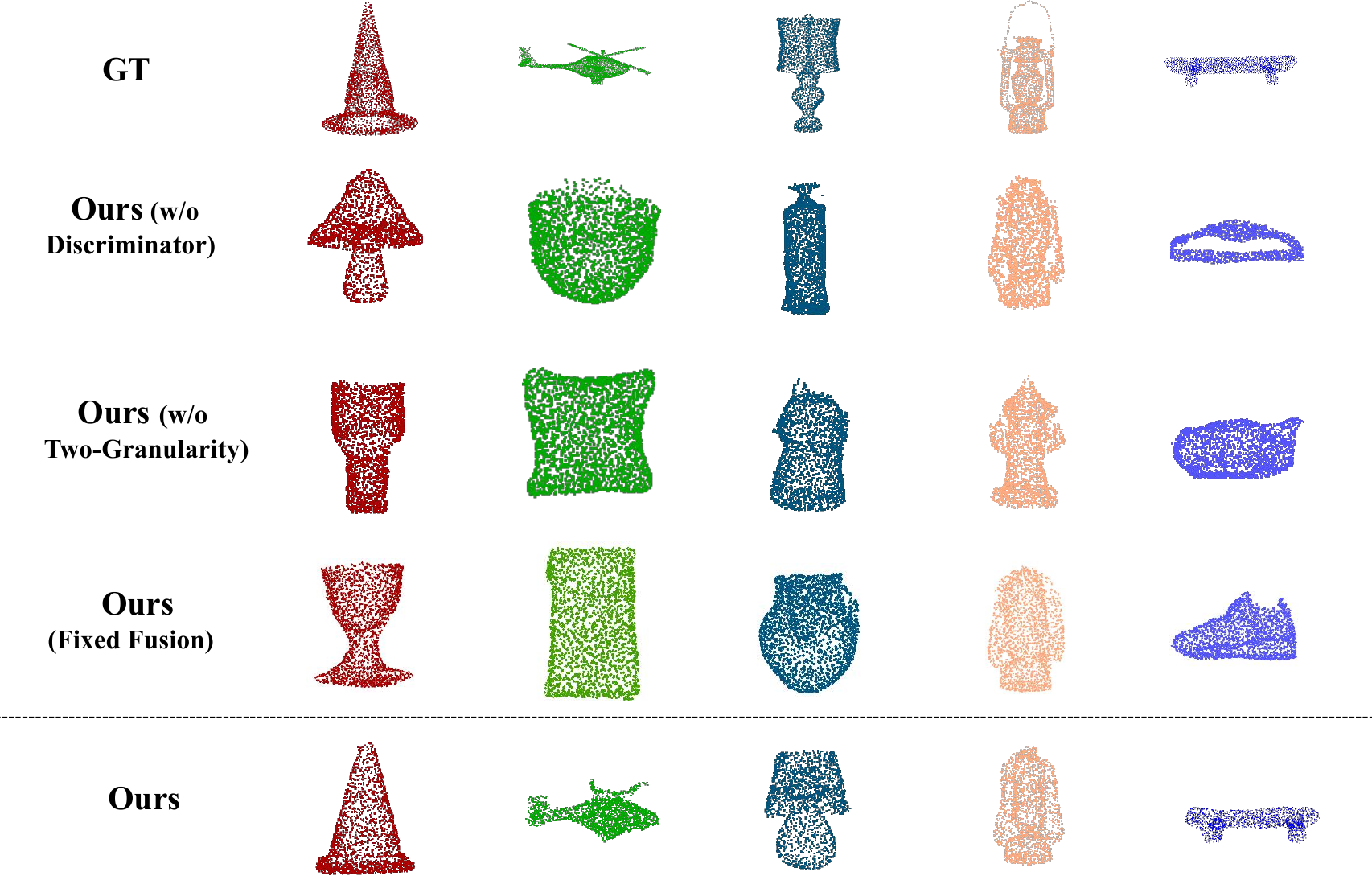}
    \caption{Qualitative ablation comparison. From top to bottom: ground truth, Mind2Cloud without the adversarial discriminator, Mind2Cloud without the two-granularity decoder, Mind2Cloud with fixed fusion, and the full model.}
    \label{fig:ablation_vis}
\end{figure}

\subsubsection{Ablation Study.}
We conduct ablation studies to evaluate the contributions of key components in our framework. As reported in Table~\ref{tab:cd_emd} and Table~\ref{tab:nway}, removing the discriminator (\textit{w/o Discriminator}) leads to increased CD/EMD scores and a drop in both average and Top-1 N-way classification accuracy, suggesting that adversarial supervision enhances both geometric fidelity and semantic alignment. To evaluate our timestep-aware design, we also test a variant with a fixed equal-weight fusion across all diffusion steps (\textit{Fixed Fusion}). As shown in Table~\ref{tab:cd_emd} and Table~\ref{tab:nway}, the fixed fusion yields suboptimal performance, indicating that our performance gains stem from time-varying global/local feature modulation rather than a static combination. Similarly, removing our two-granularity fusion design altogether (\textit{w/o Two-Granularity}) results in further degradation across all metrics. Visual comparisons in Figure~\ref{fig:ablation_vis} further reveal less complete and structurally coherent shapes in both ablated variants. These results collectively confirm that both adversarial fine-tuning and the two-granularity decoder are essential for generating accurate and semantically meaningful reconstructions.

\begin{figure*}[t]
\centering
\includegraphics[width=0.75\linewidth]{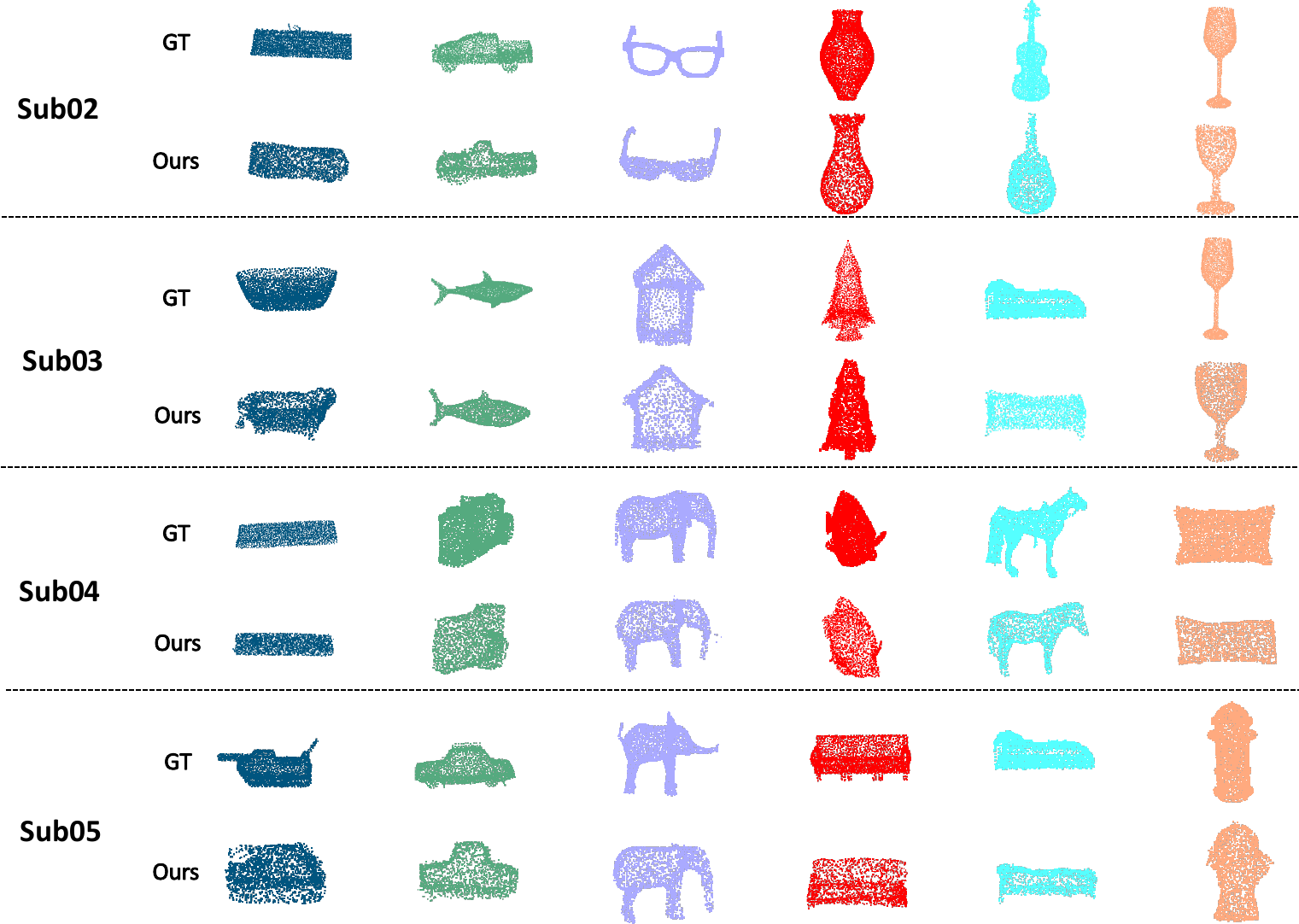} 
\caption{Subject-wise visualization results on Subject 2 to Subject 5. For each subject, the top row shows ground-truth point clouds and the bottom row shows Mind2Cloud reconstructions generated from EEG signals.}
\label{all_vis}
\end{figure*}

\subsubsection{Subject-wise Visualization Results.}

To evaluate cross-subject robustness, we present qualitative reconstructions for four subjects (Sub02–Sub05) in Figure~\ref{all_vis}. Each example shows the ground-truth point clouds and the corresponding outputs generated solely from EEG signals. Despite subject-specific variations in neural patterns, our method produces geometrically plausible and semantically consistent shapes across subjects. The reconstructed objects preserve key structural characteristics for diverse categories such as glasses, vases, animals, and furniture. Although some fine-grained surface details remain missing due to the inherently low spatial resolution of EEG signals, the results demonstrate that our framework can reliably capture meaningful 3D structure from noisy inputs.

\subsubsection{Subject-wise EEG Activation Mapping.}

To investigate inter-subject variability in cortical activity, we adopt the standard 64-channel EEG scalp partitioning into seven functional regions 

\begin{figure}[t]
    \centering
    \includegraphics[width=1.0\textwidth]{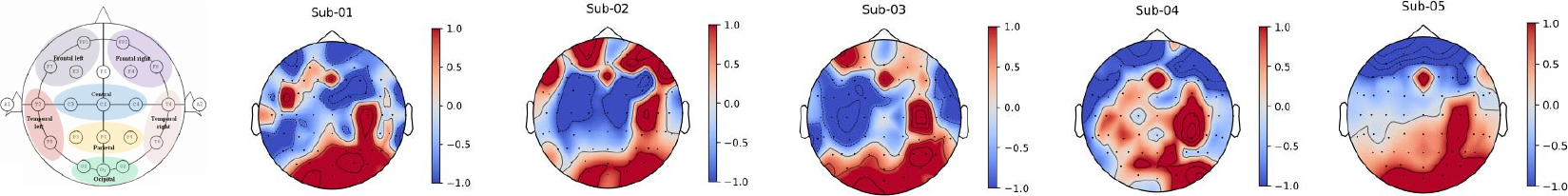}
    \caption{Normalized EEG spatial activation maps for five subjects (Sub01–Sub05). The left schematic shows the 64-channel scalp partitioning, and each subject map shows the average potential distribution across trials and time points. Red indicates higher positive voltage, while blue indicates negative or suppressed activity.}
    \label{fig:brain_regions}
\end{figure}

Figure~\ref{fig:brain_regions} (left). This schematic groups the electrodes into seven principal lobes: left/right frontal, central, parietal, left/right temporal, and occipital. Such an anatomical division enables region-specific interpretation of EEG patterns and provides a foundation for our subsequent subject-wise analysis by supporting spatial localization of neural activation.

Figure~\ref{fig:brain_regions} (right) illustrates normalized activation maps for five subjects, averaged over trials and time. Sub01 and Sub05 exhibit a center-suppressed, periphery-enhanced pattern, with diminished frontal-central activity and elevated occipital-temporal responses—consistent with their superior decoding performance (e.g., 72.22\% Top-1 for Sub01), suggesting more structured neural representations. By contrast, Sub03 and Sub04 show localized occipital-temporal activation with weaker global suppression, while Sub02 displays polarity shifts across central-frontal regions, indicating unstable engagement and moderate accuracy. These subject-specific patterns reflect inherent variability in perceptual strategies and neural encoding, reinforcing the importance of personalized modeling in EEG-based 3D reconstruction.

\section{Conclusion}
We propose an end-to-end framework that reconstructs 3D shapes from EEG signals using a two-stream EEG encoder and a time-aware two-granularity diffusion decoder. Guided by a learnable fusion strategy, the model balances global and local features throughout denoising. An adversarial point cloud discriminator further enhances geometric fidelity. While achieving strong 3D reconstruction and classification results, the model does not yet reconstruct photorealistic appearance and leaves room for stronger cross-subject generalization. Future work will focus on color reconstruction via vision-language priors and improving robustness across subjects through domain adaptation and contrastive learning, advancing semantic brain decoding and practical non-invasive BCIs. 


\section*{Acknowledgments}
This work is supported by the Guangdong Pearl River Talent Program (No. 2023QN10X721), and the Natural Science Foundation of Guangdong Province (No. 2025A1515010454).

%
%
\bibliographystyle{splncs04}
\bibliography{main}
\end{document}